\documentclass[sigconf]{acmart}

\AtBeginDocument{%
  \providecommand\BibTeX{{%
    \normalfont B\kern-0.5em{\scshape i\kern-0.25em b}\kern-0.8em\TeX}}}

\copyrightyear{2020}
\acmYear{2020}
\setcopyright{acmcopyright}\acmConference[MM '20]{Proceedings of the 28th ACM International Conference on Multimedia}{October 12--16, 2020}{Seattle, WA, USA}
\acmBooktitle{Proceedings of the 28th ACM International Conference on Multimedia (MM '20), October 12--16, 2020, Seattle, WA, USA}
\acmPrice{15.00}
\acmDOI{10.1145/3394171.3413715}
\acmISBN{978-1-4503-7988-5/20/10}

\usepackage{graphicx}
\usepackage{amsmath}
\usepackage{amsthm}
\usepackage{booktabs}
\usepackage{algorithm}
\usepackage{algorithmicx}
\usepackage{algpseudocode}

\usepackage[normalem]{ulem}
\usepackage{multirow}
\usepackage{url}
\usepackage{setspace}
\usepackage{balance}

\begin{document}
\fancyhead{}
\title{Dynamic Context-guided Capsule Network \\ for Multimodal Machine Translation}


\author{Huan Lin}
\email{huanlin@stu.xmu.edu.cn}
\affiliation{%
  \institution{Xiamen University, China}
}

\author{Fandong Meng}
\email{fandongmeng@tencent.com}
\affiliation{%
  \institution{Tencent WeChat AI, China}
}

\author{Jinsong Su}
\authornote{Corresponding author.}
\email{jssu@xmu.edu.cn}
\affiliation{%
  \institution{Xiamen University, China}
}

\author{Yongjing Yin}
\affiliation{%
  \institution{Xiamen University, China}
}

\author{Zhengyuan Yang}
\affiliation{%
  \institution{University of Rochester, USA}
}

\author{Yubin Ge}
\affiliation{%
  \institution{University of Illinois at Urbana-Champaign, USA}
}

\author{Jie Zhou}
\affiliation{%
  \institution{Tencent WeChat AI, China}
}

\author{Jiebo Luo}
\affiliation{%
  \institution{University of Rochester, USA}
}

\renewcommand{\shortauthors}{Lin, et al.}

\begin{abstract}
Multimodal machine translation (MMT), which mainly focuses on enhancing text-only translation with visual features, has attracted considerable attention from both computer vision and natural language processing communities. 
Most current MMT models resort to attention mechanism, global context modeling or multimodal joint representation learning to utilize visual features. However, the attention mechanism lacks sufficient semantic interactions between modalities while the other two provide fixed visual context, which is unsuitable for modeling the observed variability when generating translation. 
To address the above issues, in this paper, we propose a novel \textbf{D}ynamic \textbf{C}ontext-guided \textbf{C}apsule \textbf{N}etwork (DCCN) for MMT.  Specifically, at each timestep of decoding, we first employ the conventional source-target attention to produce a  timestep-specific source-side context vector. Next, DCCN takes this vector as input and uses it to guide the iterative extraction of related visual features via a context-guided dynamic routing mechanism. Particularly, we represent the input image with global and regional visual features, we introduce two parallel DCCNs to model multimodal context vectors with visual features at different granularities. Finally, we obtain two multimodal context vectors, which are fused and incorporated into the decoder for the prediction of the target word.
Experimental results on the Multi30K dataset of English-to-German and English-to-French translation demonstrate the superiority of DCCN. 
Our code is available on \url{https://github.com/DeepLearnXMU/MM-DCCN}.
\end{abstract}

\ccsdesc[500]{Computing methodologies~Natural language generation}
\ccsdesc[500]{Computing methodologies~Computer vision}
\ccsdesc[500]{Computing methodologies~Machine translation}
\keywords{Multimodal Machine Translation, Capsule Network, Transformer}


\maketitle

\section{Introduction}
\begin{figure}[t]
\centering
\includegraphics[width=1\linewidth]{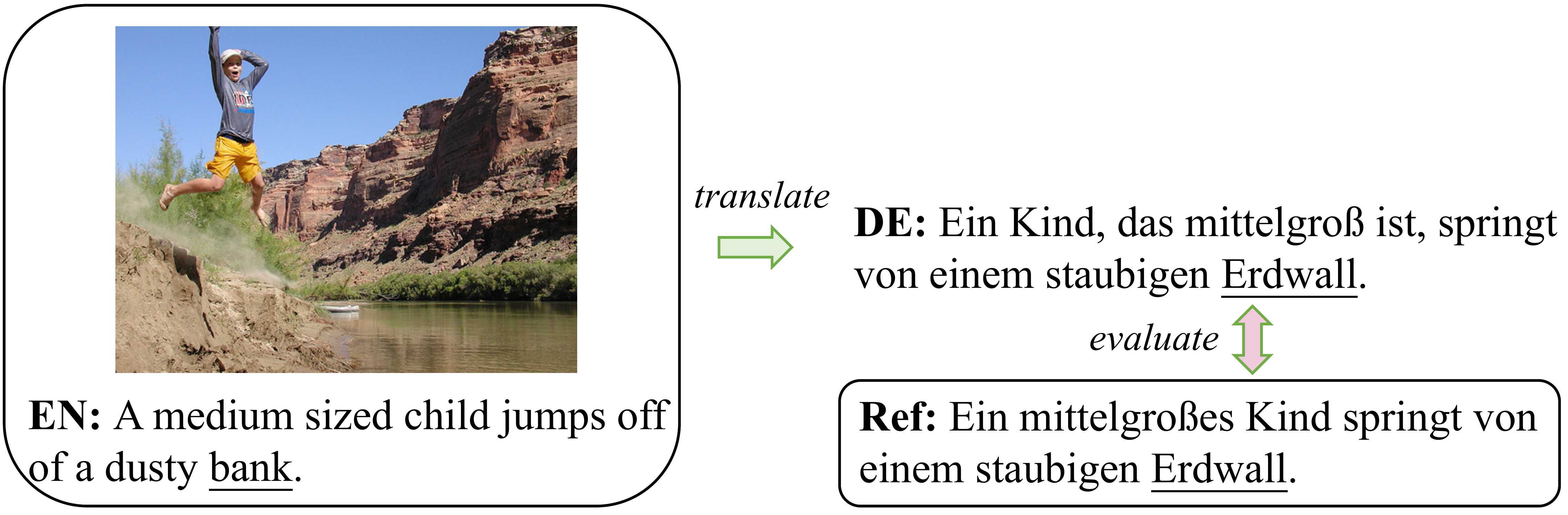} %
\caption{An example of English (EN)-to-German (DE) MMT, according to the image, \emph{``bank"} in the source sentence indicates sloping raised land rather than financial organization.} 
\label{fig_example}
\end{figure}
MMT significantly extends the conventional text-based machine translation by taking corresponding images as additional inputs \cite{mmt16,mmt17,mmt18}. 
The assumption behind this is that the translation is expected to be more accurate compared to purely text-based translation, since the visual context helps to solve data sparsity and ambiguity problems \cite{acl19:twopass}. As shown in Figure \ref{fig_example}, with the help of the image, MMT model is able to correctly translate  \emph{``bank"} to \emph{``erdwall"}. 
Overall, 
the research on MMT is of great significance. On the one hand, 
similar to other multimodal tasks such as image captioning \cite{mmcaption0,mmcaption1,mmcaption2,mmcaption3} and visual question answering \cite{mmvqa0,mmvqa1,mmvqa2,mmvqa3}, 
MMT involves computer vision and natural language processing (NLP) and proves the effectiveness of visual features in translation tasks. In other words, it not only requires an algorithm with in-depth understanding of visual contexts, but also connects its interpretation with a language model to create a natural sentence. 
On the other hand, 
MMT has wide applications, such as translating multimedia news, product information and movie subtitles \cite{zhou2018visual}. Therefore, MMT has become an attractive but challenging multimodal task.



Very importantly, 
one of the key issues in MMT is how to effectively utilize visual features during the process of translation. 
To achieve this goal, three categories of methods have been investigated: 
(1) exploiting visual features as global visual context \cite{huang2016attention,calixto2017incorporating,gronroos2018memad};
(2) applying attention mechanism to extract visual context, where 
one common approach is to employ a timestep-specific attention mechanism to extract visual context \cite{calixto2017doubly,delbrouck2017multimodal,helcl2018cuni} and 
another way is to use source hidden states to consider visual features and then use the obtained invariant visual context as a complement to source hidden states \cite{Delbrouck:NIPS17workshop,arslan2018doubly};   
(3) learning multimodal joint representations \cite{elliott2017imagination,zhou2018visual,calixto2019latent}. 
Despite their successes, 
these approaches still have various shortcomings. 
First, 
global visual context and learning multimodal joint representations can not encode the observed variability when generating translation. Second, 
according to previous studies \cite{lu2016hierarchical,wu2018you}, extracting visual context is beyond the capacity of a single-step attention due to the complexity of multimodal tasks. 
Although multiple attention layers can refine the extraction of visual context, the improvement is still limited. One possible reason is that too many parameters of multiple attention layers make the model vulnerable to over-fitting, especially when limited training examples are given in MMT. 
Moreover, the above methods only use visual features at global or regional level, which is unable to offer sufficient visual guidance. 
As a result, 
visual features are not fully utilized, limiting the potential of MMT models.

To overcome these issues, in this paper, we propose a novel \textbf{D}ynamic \textbf{C}ontext-guided \textbf{C}apsule \textbf{N}etwork (DCCN) for MMT. 
At each timestep of decoding, 
we first employ the standard source-target attention to produce a timestep-specific source-side context vector. 
Next, 
DCCN takes this context vector as input and uses it to guide the iterative extraction of related visual context during the dynamic routing process, 
where a multimodal context vector is updated simultaneously. 
In particular, to fully exploit image information, we employ DCCN to extract visual features at  two complementary granularities: global visual features and regional visual features, respectively. 
In this way, we can obtain two multimodal context vectors that are then fused for the prediction of the current target word. Compared with previous studies, 
DCCN is able to dynamically extract visual context without introducing a large number of parameters, which is suitable to  model such kind of variability observed in machine translation. 
Potentially, DCCN learns a better multimodal joint representation for MMT. Therefore, it is also applicable to other related tasks that require a joint representation of two different modalities, such as visual question answering.
In summary, 
the major contributions of our work are three-fold: 
\begin{itemize}
\item 
We introduce a capsule network to effectively capture visual features at different granularities for MMT, 
which has advantage of effectively capturing visual features without explosive parameters. To the best of our knowledge, our work is the first attempt to explore a capsule network to extract visual features for MMT.

\item
We propose a novel context-guided dynamic routing for the capsule network,
which uses the timestep-specific source-side context vector as the guiding signal to dynamically produce a multimodal context vector for MMT. 

\item 
We conduct experiments on the Multi30K English-German and English-French datasets. Experimental results show that our model significantly outperforms several competitive MMT models.
\end{itemize}

\begin{figure*}[h]
\centering
\includegraphics[width=0.95\textwidth]{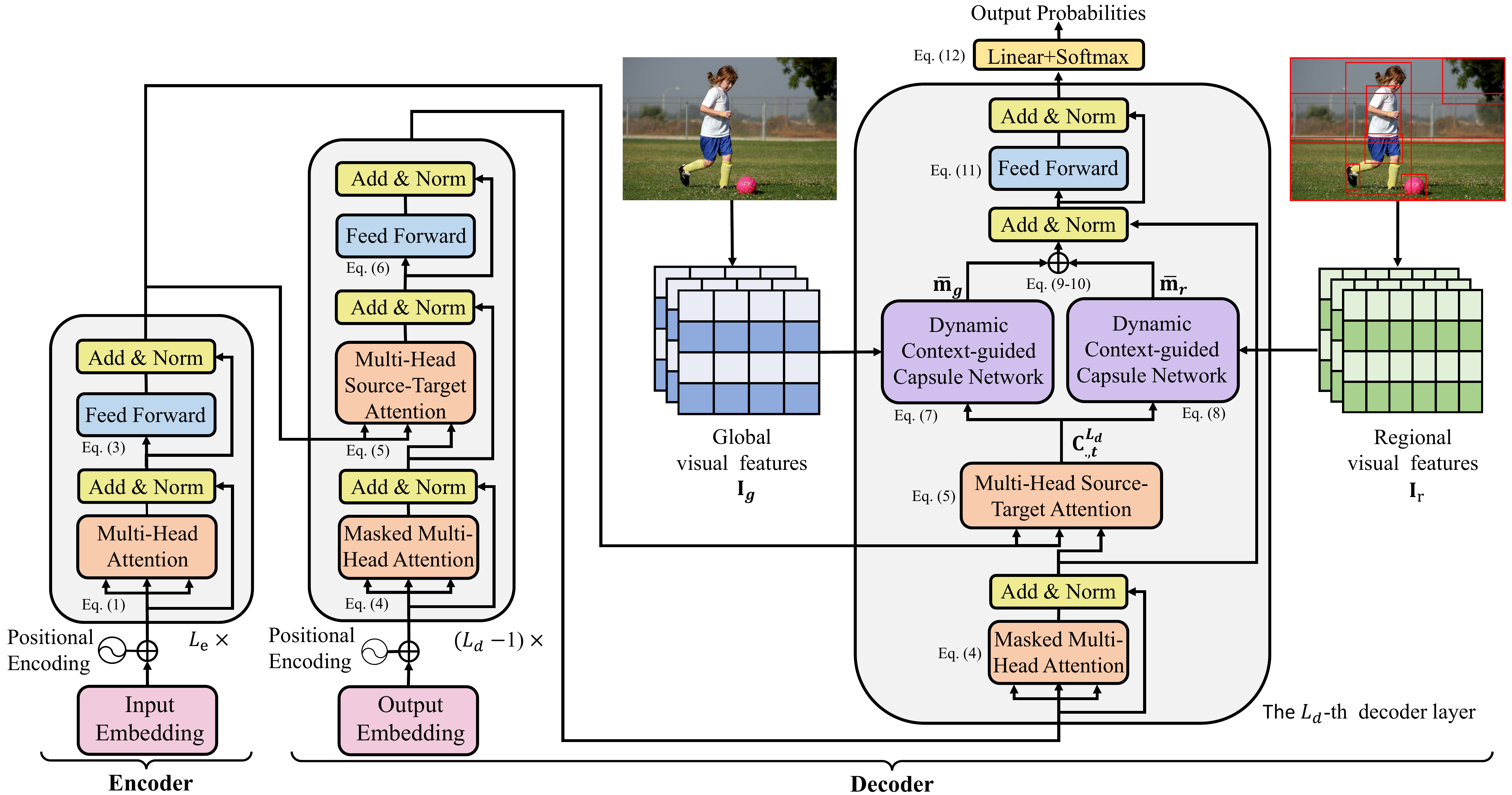} %
\caption{The architecture of our model. 
Note that we only equip the last decoder layer with two DCCNs. Using the timestep-specific source-side context vector $\textbf{C}^{(L_d)}_{.,t}$ as guidance, these two DCCNs iteratively extract global and regional visual features to form two multimodal context vectors: $\mathbf{\overline{m}}_{g}$ and $\mathbf{\overline{m}}_{r}$, respectively.
}
\label{fig1}
\end{figure*}

\section{Related Work}
The related work mainly includes the studies of multimodal context modeling in MMT and capsule networks. 

\textbf{Multimodal context modeling in MMT}. How to fully exploit context for neural machine translation has always been a hot research topic \cite{vnmt2016,context17,context18,taslp2018,vrnmt2018,abd2018,emnlp2018,ai2019,tpami2019}, which is the same for MMT. 
The commonly-used approaches to extract multimodal context for MMT can be classified into three categories:
(1) Learning global visual features for MMT. 
For instance, 
\citeauthor{huang2016attention}~\shortcite{huang2016attention} concatenated global and regional visual features with source sequences.  \citeauthor{calixto2017incorporating}~\shortcite{calixto2017incorporating} utilized global visual features as additional tokens in the source sequence, to initialize the encoder hidden states or initialize the first decoder hidden state. 
(2) Leveraging attention mechanism to exploit visual features.
In this aspect,
\citeauthor{caglayan2016multimodal}~\shortcite{caglayan2016multimodal} and \citeauthor{calixto2017doubly}\shortcite{calixto2017doubly} incorporated spatial visual features into the MMT model via an independent attention mechanism. 
Furthermore,  
\citeauthor{delbrouck2017multimodal}~\shortcite{delbrouck2017multimodal} employed Compact Bilinear Pooling to fuse the attention-based context vectors of two modalities. 
Meanwhile,  
\citeauthor{libovicky2017attention}~\shortcite{libovicky2017attention} explored flat and hierarchical combinations to fuse the attention-based context vectors of two modalities. 
Instead of using attention mechanism, \citeauthor{gronroos2018memad}~\shortcite{gronroos2018memad} introduced a gating layer to modify the prediction distribution on both visual features and decoder states. 
Unlike previous studies, \citeauthor{Delbrouck:NIPS17workshop}~\shortcite{Delbrouck:NIPS17workshop} utilized the attention mechanism on visual inputs for the source hidden states.
Along this line, \citeauthor{arslan2018doubly}~\shortcite{arslan2018doubly} extended this approach into Transformer,
and
\citeauthor{helcl2018cuni}~\shortcite{helcl2018cuni}
used timestep-specific source-side context vector as attention query to dynamically produce the visual context vectors. 
(3) Applying multi-task learning to jointly model translation task with other visual related tasks. 
For example, 
\citeauthor{elliott2017imagination}~\shortcite{elliott2017imagination} decomposed multimodal translation into two sub-tasks: 
learning to translate and learning visually grounded representations. \citeauthor{zhou2018visual}~\shortcite{zhou2018visual} optimized the learning of a shared visual language embedding and a multimodal attention-based translator. 
\citeauthor{calixto2019latent}~\shortcite{calixto2019latent} introduced a continuous latent variable for MMT, 
which contains the underlying semantic information extracted from texts and images.
Recently, \citeauthor{yyj}~\shortcite{yyj} uses a unified multi-modal graph to  capture various semantic relationships between multi-modal semantic units.

\textbf{Capsule Network}.
Recently, capsule network has been widely used in computer vision 
\cite{DBLP:conf/iclr/XinyiC19,DBLP:conf/eccv/LiGDOW18,DBLP:conf/iccv/SinghN0V19,DBLP:journals/spl/XiangZTZX18,DBLP:conf/eccv/JaiswalA0N18,DBLP:journals/corr/abs-1812-00303} 
and NLP \cite{DBLP:conf/iclr/XinyiC19,chen2019transfer,yang-etal-2018-investigating,aly2019hierarchical} tasks. 
Specific to machine translation, \citeauthor{DBLP:conf/emnlp/Wang19}~\shortcite{DBLP:conf/emnlp/Wang19} employed dynamic routing algorithm to model child-parent relationships between lower and higher encoder layers. 
\citeauthor{DBLP:conf/emnlp/YangZMGFZ19}~\shortcite{DBLP:conf/emnlp/YangZMGFZ19} proposed a query-guided capsule networks to cluster context information into  different perspectives from which the target translation may concern. 
\citeauthor{DBLP:conf/emnlp/ZhengHTDC19}~\shortcite{DBLP:conf/emnlp/ZhengHTDC19} separated translated and untranslated source words into different groups of capsules. 

To the best of our knowledge, 
our work is the first attempt to introduce the capsule network into MMT. 
Furthermore, 
we use the timestep-specific source-side context vector rather than static source hidden states to guide the routing procedure. 
Therefore, we can dynamically produce visual context for MMT.

\section{Our Model}
As shown in Figure \ref{fig1}, our model is based on Transformer \cite{vaswani2017attention}. 
The most important feature of our model is that two DCCNs are introduced to dynamically learn multimodal context vectors for  generating translations.


\subsection{Encoder}

Given the source sentence $X$, 
we represent each source word as the sum of its word embedding and positional encoding. 
Next, 
we follow \citeauthor{vaswani2017attention}~\shortcite{vaswani2017attention} to use a stack of $L_{e}$ identical layers to encode $X$, 
where each layer consists of two sub-layers. 
Note that we also introduce residual connection and layer normalization to each sub-layer, of which the descriptions are omitted.

Specifically,
at the $l$-th layer ($1$$\leq$$l$$\leq$$L_{e}$),
the first sub-layer is a multi-head self-attention: 
\begin{equation} 
\mathbf{H}^{(l)}_e = \mathrm{MultiHead}(\mathbf{S}^{(l-1)}, \mathbf{S}^{(l-1)}, \mathbf{S}^{(l-1)}), 
\end{equation} 
where $\mathbf{H}^{(l)}_e$ is the temporary encoder hidden states, 
$\mathrm{MultiHead}(*)$ is a multi-head self-attention function, 
and $\mathbf{S}^{(l-1)}$$\in$$\mathbb{R}^{d_w\times |X|}$ is the representation of the source sentence at the ($l$-1)-th layer, $d_w$ is the model dimension.
Particularly,
$\mathbf{S}^{(0)}$ is the concatenation of all source word embeddings. 
Taking a query matrix $\mathbf{Q}$, a key matrix $\mathbf{K}$ and a value matrix $\mathbf{V}$ as inputs, 
$\mathrm{MultiHead}(*)$ is defined as follows:

\begin{align}
\mathrm{MultiHead}(\mathbf{Q},\mathbf{K},\mathbf{V})&=\mathrm{Concat}({head}_1,...,{head}_{N_h})\mathbf{W}^C \\
where~{head}_k&=\mathrm{Attention}(\mathbf{QW}_k^Q,\mathbf{KW}_k^K,\mathbf{VW}_k^V) \notag \\
&=\mathrm{Softmax}(\frac{\mathbf{QW}_k^Q(\mathbf{KW}_k^K)^T}{\sqrt{d_w}})\mathbf{VW}_k^V, \notag
\end{align}
where $\mathbf{W}^C$ and $\mathbf{W}^*_k$ are learnable parameter matrices.

The second sub-layer is a position-wise fully connected feed-forward network. It is applied to each position separately and identically,
forming the representation $\mathbf{S}^{(l)}$ of the source sentence as
\begin{equation}
\mathbf{S}^{(l)} = \mathrm{FFN}(\textbf{H}^{(l)}_e),
\end{equation}
where $\mathrm{FFN}(*)$ is a position-wise feed-forward function.

\subsection{Decoder}

As shown in Figure \ref{fig1},
our decoder is an extension of the Transformer decoder \cite{vaswani2017attention}.
It takes the already generated sequence as inputs and uses a stack of $L_d$ identical layers to produce target-side hidden states.
Similar to the standard Transformer decoder, each layer of our decoder contains three sub-layers. 
The only difference is that at the last decoder layer,  
two DCCNs are equipped to produce timestep-specific multimodal context vectors for MMT.

Specifically, 
the first sub-layer is also a multi-head self-attention:
\begin{align}
\mathbf{H}^{(l)}_d = \mathrm{MultiHead}(\mathbf{T}^{(l-1)}, \mathbf{T}^{(l-1)}, \mathbf{T}^{(l-1)}), 1\leq{l}\leq{L_d},
\end{align}
where $\mathbf{H}^{(l)}_d$ denotes the temporary decoder hidden states, produced by a multi-head self-attention mechanism fed with the target-side hidden states $\mathbf{T}^{(l-1)}$ at the previous layer.

Typically,
at the second sub-layer,
a multi-head source-target attention mechanism is used to dynamically produce the timestep-specific source-side context vectors $\mathbf{C}^{(l)}$:
\begin{align}
\mathbf{C}^{(l)} = \mathrm{MultiHead}(\mathbf{H}^{(l)}_d, \mathbf{S}^{(L_{e})}, \mathbf{S}^{(L_{e})}), 1\leq{l}\leq{L_d}.
\end{align}

The third sub-layer is 
a position-wise fully connected feed-forward neural network, of which the definition depends on the decoder layer. At the first $(L_d-1)$ layers, this sub-layer produces the target-side hidden states $\mathbf{T}^{(l)}$ as follows:
\begin{align}
\mathbf{T}^{(l)} = \mathrm{FFN}(\mathbf{C}^{(l)}), 1\leq{l}\leq{L_d}-1. \label{equ_ffn}
\end{align}
Very importantly, at the last ($L_d$-th) decoder layer, 
we introduce two DCCNs between the second and third sub-layers to  learn multimodal context representations at the $t$-th timestep as follows: 

\begin{align}
\mathbf{\overline{m}}_{g} = \mathrm{CapsuleNet}(\mathbf{C}_{.,t}^{(L_d)},\mathbf{I}_g), \\
\mathbf{\overline{m}}_{r} = \mathrm{CapsuleNet}(\mathbf{C}_{.,t}^{(L_d)},\mathbf{I}_r),
\end{align}
where $\mathrm{CapsuleNet}(*)$ is a context-guided dynamic routing function, $\mathbf{C}_{.,t}^{(L_d)}$ is the $t$-th column vector of $\mathbf{C}^{(L_d)}$, representing the source-side context vectors at the $t$-th timestep,  $\mathbf{I}_g$ and $\mathbf{I}_r$ represent global visual features and local visual features respectively, as depicted in subsection \ref{sec_visual_features}. 
By using DCCN, we can iteratively extract the related visual features to dynamically produce better multimodal context representations for MMT. We will describe this procedure in subsection \ref{sec_dccn}. 
Next, we fuse $\mathbf{\overline{m}}_{g}$ and $\mathbf{\overline{m}}_{r}$ via the following gating mechanism: 
\begin{align}
&\mathbf{M}_{.,t}^{(L_d)} = \mathbf{\alpha}\mathbf{\overline{m}}_{g} + (1-\mathbf{\alpha})\mathbf{\overline{m}}_{r} \\
&\mathbf{\alpha} = \text{Sigmoid}(\mathbf{W}_g\mathbf{\overline{m}}_{g}+\mathbf{W}_r\mathbf{\overline{m}}_{r}),
\end{align}
where $\mathbf{W}_{g}$ and $\mathbf{W}_r$ are learnable parameters. 
Correspondingly, the \textbf{Eq.} \ref{equ_ffn} at the last decoder layer becomes 
\begin{align}
\mathbf{T}^{(L_d)} = \mathrm{FFN}(\mathbf{M}^{(L_d)}). \label{equ_ffn2}
\end{align}

Finally, with the target-side hidden states generated by \textbf{Eq.} \ref{equ_ffn2}, 
our decoder adopts a Softmax layer to generate the probability distribution of the current target word $y_{t}$:
\begin{align}
P(y_{t}|X,Y_{<t};\theta)\propto\mathrm{exp}(\mathbf{WT}^{(L_{d})}_{.,t}),
\end{align}
where $Y_{<t}$ is the previously generated target words ${y_1}{y_2}\dots{y_{t-1}}$, $W$$\in$$\mathbb{R}^{|V_{y}|\times d_w}$ is a model parameter, $V_{y}$ is the target vocabulary, 
and $\mathbf{T}^{(L_{d})}_{.,t}$ is the $t$-th column vector of $\mathbf{T}^{(l)}$ for predicting $y_t$.

\subsubsection{Visual Features}\label{sec_visual_features}
To fully exploit visual information for MMT, we investigate two kinds of visual features to enhance text-based translation: 
(1) \textbf{global visual features}, which represent an input image with high-level concepts. Here we use the \textit{res4f} layer activations of pre-trained 50-layer Residual Network (ResNet-50)\cite{resnet} as global visual features. These spatial features encode an image in a 14×14 grid, where each grid is represented by a 1,024D feature vector, only encoding the information about the specific part of the image. 
Before fed into DCCN, we first follow \citeauthor{calixto2017doubly}~\shortcite{calixto2017doubly} to transform global visual features into a 196×256 matrix $\mathbf{I}_g$ 
where each of the 196 rows consists of a 256D feature vector; and 
(2) \textbf{regional visual features} that illustrate class annotations of each region (e.g. \emph{cat, arms, peak}). 
Following \citeauthor{bottom-up-attention}~\shortcite{bottom-up-attention}, we employ the R-CNN based bottom-up attention to identify the regions with class annotations. 
For each region, we generate the corresponding prediction probability distribution over 1,600 classes from Visual Genome. 
To represent each region as a vector, we project its class annotations into word embeddings and define the region vector as the weighted sum of its class annotation embeddings. Finally, all region vectors are concatenated to represent the semantics of input image. In practice, we keep the number of predicted regions up to 10 so as to reduce negative effects of abundant regions, therefore the regional visual features can be represented as a 10×256 matrix $\mathbf{I}_r$,  
where each of the 10 rows consists of a 256D feature vector.

\subsubsection{Dynamic Context-guided Capsule Network}\label{sec_dccn}

\begin{figure}[t]
\centering
\includegraphics[width=0.8\linewidth]{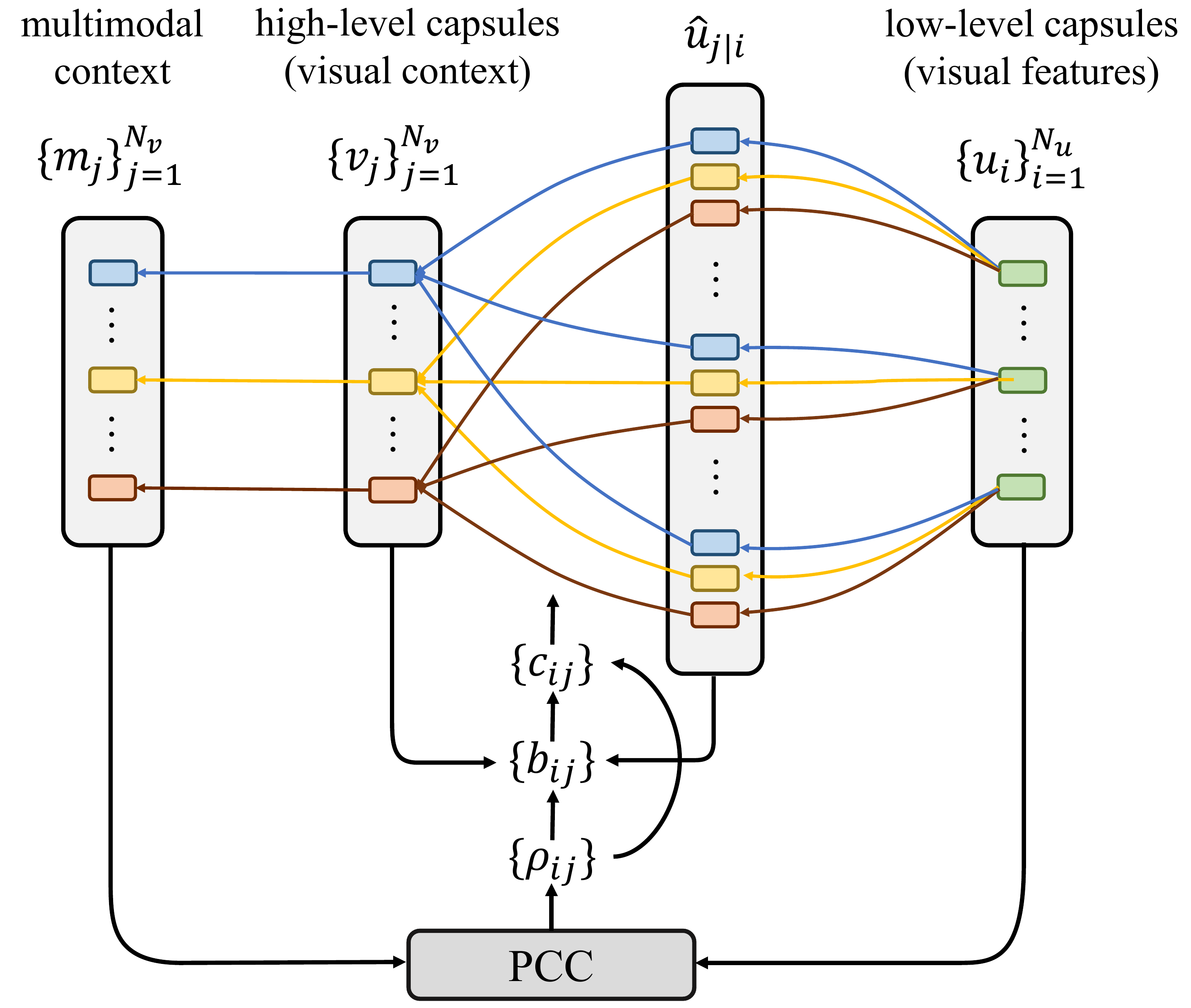} 
\caption{The routing procedure of DCCN}
\label{Fig_capsmall}
\end{figure}

Our DCCN is a significant extension of the conventional capsule network. 
Thus, it retains the advantages on iterative feature extraction of capsule network, 
which has shown effective in many computer vision  \cite{DBLP:conf/iclr/XinyiC19,DBLP:conf/eccv/LiGDOW18,DBLP:conf/iccv/SinghN0V19,DBLP:journals/spl/XiangZTZX18,DBLP:conf/eccv/JaiswalA0N18,DBLP:journals/corr/abs-1812-00303} and NLP \cite{DBLP:conf/iclr/XinyiC19,chen2019transfer,yang-etal-2018-investigating,aly2019hierarchical,DBLP:conf/emnlp/YangZMGFZ19,DBLP:conf/emnlp/ZhengHTDC19,DBLP:conf/emnlp/Wang19} tasks. 
More importantly, unlike the conventional capsule network that only captures static visual context, DCCN introduces the timestep-specific source-side context vector to guide the extraction of multimodal context, which can model the observed variability during translation.

Figure \ref{Fig_capsmall} shows the architecture of DCCN. 
Similar to the conventional capsule network, DCCN consists of 
(1) low-level capsules $\{\mathbf{u}_i\}_{i=1}^{N_u}$ (the first column from the right) encoding the visual features of the input image and 
(2) high-level capsules $\{\mathbf{v}_j\}_{j=1}^{N_v}$ (the second column from the left) encoding the related visual features. 
Besides, it includes multimodal context capsules $\{\mathbf{m}_j\}_{j=1}^{N_v}$ (the first column from the left), where $\mathbf{m}_j$ indicates a temporary multimodal context vector iteratively updated with $\mathbf{v}_j$ and is used as query to guide the extraction of related visual features. 

\begin{spacing}{0.8}
\begin{algorithm}[t!]
\caption{Context-guided Dynamic Routing Mechanism.} \label{routing}
\begin{algorithmic}[1]
\State \textbf{Input:} 
the timestep-specific source-side context vector $\mathbf{C}_{.,t}^{(L_d)}$, 
the input image $\mathbf{I}$, 
the low-level capsule number $N_u$,
the high-level capsule number $N_v$, 
the iteration number $N_{itr}$
\State \textbf{Output:} the multimodal context vector $\mathbf{\overline{m}}$

\For{$i=1\dots N_u$}
\State Initialize the low-level capsules $\mathbf{u}_i$ with $\mathbf{I}_i$ \label{line_ui}
\EndFor
\For{$j=1\dots N_v$}
\State Initialize the multimodal context capsules $\mathbf{m}_j$ with $\mathbf{C}_{.,t}^{(L_d)}$ \label{line_cs}
\EndFor

\For{each low-level capsule $\mathbf{u}_i$}
\For{each high-level capsule $\mathbf{v}_j$}
\State $\mathbf{b}_{ij} \gets 0$
\State $\mathbf{\hat{u}}_{j|i} = \mathbf{W}_{ij}\mathbf{u}_i$ \label{line_uhij}
\State $\mathbf{\rho}_{ij} \gets \mathrm{tanh}(\mathrm{PCC}(\mathbf{u}_i, \mathbf{W}_m\mathbf{m}_j))$  \label{line_pij}
\EndFor
\EndFor

\For{$itr = 1 \dots N_{itr}$} \label{line_begin}

\For{each low-level capsule $\mathbf{u}_i$}
\For{each high-level capsule $\mathbf{v}_j$}
\State $\{\mathbf{c}_{ij}\}_{j=1}^{N_v} \gets \mathrm{Softmax}(\{\mathbf{b}_{ij}\}_{j=1}^{N_v})$ \label{line_ci}
\EndFor
\EndFor

\For{each high-level capsule $\mathbf{v}_j$}
\State $\mathbf{v}_j \gets \sum_i (\mathbf{c}_{ij}+\mathbf{\rho}_{ij}){\mathbf{\hat u}}_{j|i}$ \label{line_vj}
\State $\mathbf{m}_j \gets \mathbf{m}_j \odot \mathbf{W}_v\mathbf{v}_j$ \label{line_mj}
\EndFor

\For{each low-level capsule $\mathbf{u}_i$}
\For{each high-level capsule $\mathbf{v}_j$}
\State $\rho_{ij} \gets \mathrm{tanh}(\mathrm{PCC}(\mathbf{u}_i, \mathbf{W}_m\mathbf{m}_j))$   \label{line_pij2}
\State $\mathbf{b}_{ij} \gets \mathbf{b}_{ij} + \rho_{ij}(\mathbf{\hat{u}}_{j|i} \cdot \mathbf{v}_j)$ \label{line_bij} 
\EndFor
\EndFor

\EndFor \label{line_end}
\State $\mathbf{\overline{m}}=\mathrm{FuseMultimodalContext}(\mathbf{m}_1,\dots,\mathbf{m}_{N_v})$ \label{line_cm}
\end{algorithmic}
\end{algorithm}
\end{spacing}

We summarize the training procedure in Algorithm \ref{routing}. 
Given a visual features matrix $\mathbf{I}$, 
we use low-level capsules $\{\mathbf{u}_i\}_{i=1}^{N_u}~(\mathbf{u}_i\in\mathbb{R}^{256})$ to individually encode the semantic representation of each row of  $\mathbf{I}$ (\textbf{Line} \ref{line_ui}), 
and initialize each multimodal context capsule $\mathbf{m}_j$ with the timestep-specific source-side context vector $\mathbf{C}_{.,t}^{(L_d)}$ (\textbf{Line} \ref{line_cs}). 
Following the conventional capsule network,
we introduce the matrix $\mathbf{W_{ij}}$ to transform the $i$-th low-level capsule $\mathbf{u}_i$ into $\mathbf{\hat{u}_{j|i}}$ (\textbf{Line} \ref{line_uhij}), 
which will be used to generate the $j$-th high-level capsule $\mathbf{v}_j$.

Next, we introduce the coefficient $\mathbf{\rho}_{ij}$ to measure the cross-modal correlation between $\mathbf{u}_i$ and the multimodal context vector $\mathbf{m}_j$ (\textbf{Line} \ref{line_pij}), which can be subsequently used to generate high-level capsules $\mathbf{v}_j$ and update $\mathbf{b}_{ij}$, benefiting the extraction of related visual features. 
Formally, 
the correlation function is defined as
\begin{align}
\mathbf{\rho}_{ij}=\mathrm{tanh}(\mathrm{PCC}(\mathbf{u}_i,\mathbf{W}_m\mathbf{m}_j))=\mathrm{tanh}(\frac{\text{Cov}(\mathbf{u}_i,\mathbf{W}_m\mathbf{m}_j)}{\sigma(\mathbf{u}_i) ~\sigma(\mathbf{W}_m\mathbf{m}_j)}), \label{pcc}
\end{align}
where $\text{PCC}(*)$ indicates the Pearson Correlation Coefficients, $\mathbf{W}_m$ is a parameter matrix that maps $\mathbf{m}_j$ to the same semantic space of $\mathbf{u}_i$, $\text{Cov}(*)$ is the covariance and $\sigma(*)$ is the standard deviation. 
When $\mathbf{\rho}_{ij}$ is close to +1, 
the visual features encoded by $\mathbf{u}_i$ is closely related to $\mathbf{m}_j$, 
otherwise it indicates negative correlation. 

Then, 
we conduct $N_{itr}$ iterations of routing to capture related visual features at current timestep (\textbf{Line} \ref{line_begin} to \textbf{Line} \ref{line_end}).
At each iteration,
we generate high-level capsules from low-level capsules.
To do this,
we employ a Softmax function along columns with the logits $\mathbf{b}_{ij}$ initialized as 0 to calculate the coupling coefficient $\mathbf{c}_{ij}$ (\textbf{Line} \ref{line_ci}). 
Afterwards, we generate the high-level capsule $\mathbf{v}_j$ to represent visual context as the weighted sum of $\mathbf{\hat u}_{j|i}$ according to their corresponding $\mathbf{c}_{ij}$ and the cross-modal correlation coefficient $\mathbf{\rho}_{ij}$ (\textbf{Line} \ref{line_vj}). 
Note that unlike the conventional dynamic routing algorithm \cite{sabour2017dynamic} where $\mathbf{v}_{j}$ only depends on $\mathbf{c}_{ij}$ and $\mathbf{\hat{u}_{j|i}}$, 
we further introduce $\mathbf{\rho}_{ij}$ that enables the most relevant visual features to be iteratively clustered into high-level capsules.
Note that the conventional capsule network \cite{sabour2017dynamic} uses the norm of high-level capsule to represent prediction probability, thus the norm of $\mathbf{v}_j$ is adjusted to [0,1] using the ``\emph{squashing}" function.   
Different from that, $\mathbf{v}_j$ represents visual context in DCCN. Therefore we do not apply ``\emph{squashing}" function in our algorithm. 
Further, we introduce a transformation matrix $\mathbf{W}_v$ to map visual context $\mathbf{v}_j$ into the semantic space of multimodal context and follow \citeauthor{wu2019faithful}~\shortcite{wu2019faithful} to update $\mathbf{m}_j$ with the captured visual context $\mathbf{v}_j$ (\textbf{Line} \ref{line_mj}). By doing so, 
we expect the updated multimodal context capsule can be better exploited to guide the routing procedure at the next iteration. 

Finally, we update $\mathbf{\rho}_{ij}$ (\textbf{Line }\ref{line_pij2}), and then use it to guide the updating of $\mathbf{b}_{ij}$ (\textbf{Line }\ref{line_bij}). Different from conventional dynamic routing algorithm, where $\mathbf{b}_{ij}$ only depends on the cumulative ``agreement'' between $\mathbf{u}_i$ and $\mathbf{v}_j$, we control the updating range of $\mathbf{b}_{ij}$ according to the cross-modal correlation $\mathbf{\rho}_{ij}$.

Through $N_{itr}$ iterations of routing, we obtain $N_{v}$ multimodal context capsules  $\{\mathbf{m}_j\}^{N_{v}}_{j=1}$,  fused by a linear transformation to produce the final multimodal context vector $\mathbf{\overline{m}}$ (\textbf{Line} \ref{line_cm}). 

\section{Experiments}
\subsection{Dataset}
To investigate the effectiveness of our proposed model, we  conduct experiments on the Multi30K dataset \cite{elliott2016multi30k}, 
which is an extended version of the Flickr30K Entities and has been widely used in MMT \cite{mmt18,mmt17,mmt16}.
For each image, one of the English (EN) descriptions was selected and manually translated into German (DE) and French (FR) by professional translators \cite{mmt16}. The dataset
contains 29,000 instances for training, 1,024 for validation and 1,000 for testing. 
We also report results on the WMT2017 test set with 1,000 instances and the MSCOCO test set containing 461 out-of-domain instances with ambiguous verbs.
Besides, as mentioned in subsection \ref{sec_visual_features}, we represent the input image with visual features in two granularities.

We apply the MOSES scripts\footnote{http://www.statmt.org/moses/} to preprocess datasets.  We then employ the Byte Pair Encoding \cite{sennrich2015neural} with 10,000 merging operations to convert tokens into subwords.

\begin{table*}[!t]
\centering
\footnotesize
\begin{tabular}{c|l|c|cc|cc|cc}
\hline
\multirow{3}{*}{\#} & \multirow{3}{*}{\bf Model} & \multirow{3}{*}{\bf\#Params} &\multicolumn{6}{c}{\textbf{EN$\Rightarrow$DE}}\\
\cline{4-9}
& & &\multicolumn{2}{|c|}{\textbf{Test2016}}  &\multicolumn{2}{|c|}{\textbf{Test2017}}  &\multicolumn{2}{c}{\textbf{MSCOCO}}\\
\cline{4-9}
& & &\textbf{BLEU} & \textbf{METEOR} &\textbf{BLEU} & \textbf{METEOR}  &\textbf{BLEU} & \textbf{METEOR}\\
\hline
\hline
\multicolumn{9}{c}{\emph{Existing MMT Systems}} \\
\hline
1  &Stochastic attention \cite{Delbrouck:EMNLP17}        &--   & 38.2 & 55.4  & --   & --    & --   & -- \\
2  &Imagination \cite{elliott2017imagination}            &--   & 36.8 & 55.8  & --   & --    & --   & -- \\
3  &Fusion-conv \cite{caglayan2017lium}                  &--   & 37.0 & 57.0  & 29.8 & 51.2  & 25.1 & 46.0 \\
4  &Trg-mul \cite{caglayan2017lium}                      &--   & 37.8 & 57.7  & 30.7 & 52.2  & 26.4 & 47.4 \\
5  &Latent Variable MMT \cite{calixto2019latent}         &--   & 37.7 & 56.0  & 30.1 & 49.9  & 25.5 & 44.8 \\
6  &Deliberation Network \cite{acl19:twopass}            &--   & 38.0 & 55.6  & --   & --    & --   & -- \\
\hline
\multicolumn{9}{c}{\emph{Our MMT Systems}} \\
\hline
7  &Transformer \cite{vaswani2017attention}            & 16.1M & 38.4 & 56.0   & 29.4 & 48.8    & 25.3 & 44.4 \\
8 &Encoder-attention \cite{Delbrouck:NIPS17workshop}   & +1.1M & 39.0 & 56.6   & 29.9 & 49.5      & 26.0 & 45.5 \\
9 &Doubly-attention \cite{helcl2018cuni}               & +4.0M & 38.7 & 56.4   & 30.4 & 49.1      & 25.5 & 44.7 \\	
\hline
10 &DCCN  & +1.0M 
& $\textbf{39.7}^{\ddagger*\bigtriangleup\bigtriangleup}$ & $ \textbf{56.8}^{\ddagger\bigtriangleup}$   
& $\textbf{31.0}^{\ddagger**\bigtriangleup}$ & $ \textbf{49.9}^{\ddagger*\bigtriangleup\bigtriangleup}$   
& $\textbf{26.7}^{\ddagger*\bigtriangleup\bigtriangleup}$  & $ \textbf{45.7}^{\ddagger\bigtriangleup\bigtriangleup}$  \\
\hline
\end{tabular}
\caption{
\label{Table_En2DeMainResults}
Experimental results on the EN$\Rightarrow$DE translation task in terms of BLEU and METEOR. We also calculate statistical significance. $\ddagger$/$\dagger$: significantly better than Transformer ($p$ < 0.01/0.05), **/*: significantly better than Encoder-attention ($p$ < 0.01/0.05), $\bigtriangleup\bigtriangleup$/$\bigtriangleup$: significantly better than Doubly-attention ($p$ < 0.01/0.05). Note that DCCN outperforms most of the existing models and all baselines except for lower METEOR scores than Fusion-conv and Trg-mul, of which model parameters are selected based on METEOR.
\vspace{-10pt}
}
\end{table*}

\subsection{Setup}
We develop our proposed model based on OpenNMT Transformer \cite{klein2017opennmt}.
Since the size of training corpus is small and the trained model tends to be over-fitting, 
we first perform a small grid search to obtain a set of hyper-parameters on the EN$\Rightarrow$DE validation set. Specifically, the layer numbers of both encoder and decoder are set to 4 and the number of attention heads is set to 8. Both hidden size and embedding size are set to 256. 


As implemented in \cite{vaswani2017attention}, we use the Adam optimizer with $\beta_{2}=0.998$ and scheduled learning rate to optimize various models. The learning rate is initialized as 1.
During training, each batch consists of approximately 3,700 source and target tokens. Besides, we employ the dropout strategy \cite{srivastava2014dropout} with rate 0.5 to enhance the robustness of our model. Finally, we adopt the MultEval scripts \cite{clark2011better} to evaluate the translation quality in terms of BLEU \cite{papineni2002bleu} and METEOR \cite{denkowski2011meteor}. 
Particularly, we run all models three times for each experiment and report the average results.

The context-guided dynamic routing is important for the generation of multimodal context. 
Therefore, 
we investigate the impacts of its hyper parameters on the routing mechanism: high-level capsule number $N_v$ and routing iteration number $N_{itr}$. 
To this end, 
we try different numbers of high-level capsules and routing iteration numbers to train our model: $N_v$ from 1 to 3, $N_{itr}$ from 1 to 4 on the validation set. 
We observe that $N_v$ larger than 1 and $N_{itr}$ larger than 3 do not lead to significant improvements and increase the GPU memory requirement. 
Hence, we use $N_v$=1 and $N_{itr}$=3 in all subsequent experiments.

\subsection{Baselines}
We directly refer to our MMT model as DCCN and compare it with the following commonly-used MMT baselines: 
\begin{itemize}
    \item \textbf{Transformer \cite{vaswani2017attention}}  A text-only machine translation model.
    \item \textbf{Encoder-attention \cite{Delbrouck:NIPS17workshop}.} It incorporates an encoder-based visual attention mechanism into Transformer, which uses source hidden states to consider visual features and then augment each source hidden state with its corresponding visual context. Please note that \cite{Delbrouck:NIPS17workshop} is based on RNN and we implement it on Transformer for comparability.
    \item \textbf{Doubly-attention  \cite{helcl2018cuni}.} A doubly attentive Transformer that introduces an additional visual attention sub-layer to exploit visual features. Specifically, this sub-layer is inserted between the source-target attention and feed-forward sub-layer. For visual attention, the context vectors from the source-target attention are used as queries, and the context vectors of visual attention are fed into the feed-forward sub-layer.
\end{itemize}

We also display the performance of several dominant MMT models on the same datasets.
\textbf{Stochastic attention} \cite{Delbrouck:EMNLP17} is a stochastic and sampling-based attention mechanism, 
which focuses on only one spatial location of the image at every timestep. It is also the model of best performance in \cite{Delbrouck:EMNLP17}. 
\textbf{Imagination} \cite{elliott2017imagination} employs multitask learning to jointly two sub-tasks: translating and visually grounded representation prediction. 
\textbf{Fusion-conv} \cite{caglayan2017lium} employs a single feed-forward network to establish the attention alignment between visual features and target-side hidden states at each timestep, where all spatial locations of image are considered to derive the context vector. 
\textbf{Trg-mul} \cite{caglayan2017lium} modulates each target word embedding with visual features using element-wise multiplication.
\textbf{Latent Variable MMT} \cite{calixto2019latent} exploits the interactions between visual and textual features for MMT through a latent variable, which can be seen as a multimodal stochastic embedding of an image and its target language description. 
\textbf{Deliberation Network} \cite{acl19:twopass} is based on a translate-and-refine strategy, where visual features are only used by the decoder at the second stage.




\subsection{Results on the EN$\Rightarrow$DE Translation Task.}

\textbf{Parameters} Introducing visual feature features into Transformer model brings more parameters. As shown in Table \ref{Table_En2DeMainResults}, Transformer (\textbf{Row} 7) has 16.1M parameters. Encoder-attention (\textbf{Row} 8) adds a layer normalization, a fully connected layer and two multi-head attention layers, introducing 1.1M parameters, and Doubly attention (\textbf{Row} 9) increases 4.0M parameters by adding two layer normalization and two multi-head attention layers. By contrast, DCCN (\textbf{Row} 10) only introduces 1.0M extra parameters. 
Thus, DCCN introduces a small number of extra parameters compared to Transformer, and requires smaller parameters than two multimodal baselines. 

\begin{figure*}[!t]
\centering
\includegraphics[width=1\linewidth]{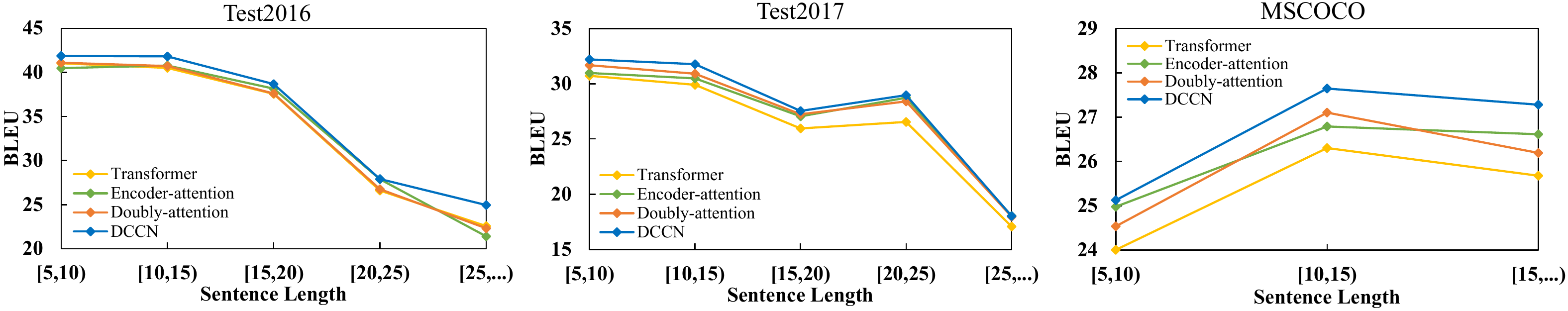}
\caption{
\label{Fig_En2De_LengthEffect}
BLEU scores on different translation groups divided according to source sentence lengths. Since MSCOCO only contains two sentences longer than 20, we divide all sentences longer than 20 into one group for testing. 
}
\end{figure*}

\begin{table*}
\centering
\footnotesize
\begin{tabular}{c|l|cc|cc|cc}  
\hline
\multirow{2}{*}{\#} &\multirow{2}{*}{\bf Model} & \multicolumn{2}{|c|}{\textbf{Test2016}} & \multicolumn{2}{|c|}{\textbf{Test2017}} & \multicolumn{2}{c}{\textbf{MSCOCO}} \\ 
\cline{3-8}
& & \textbf{BLEU} & \textbf{METEOR} & \textbf{BLEU} & \textbf{METEOR}  & \textbf{BLEU} & \textbf{METEOR}\\
\hline
\hline
1 &DCCN & 39.7 & 56.8  & 31.0 & 49.9  & 26.7 & 45.7 \\
\hline
2 & \ \ \ \ \ \ dynamic routing (global)   & 39.2 & 56.5  & 30.1 & 49.1   & 25.6 & 44.3  \\
3 & \ \ \ \ \ \ dynamic routing (regional) & 39.1 & 56.6  & 29.9 & 48.8   & 26.1 & 44.8  \\
\hline
4 &\ \ \ \ \ \ dynamic routing (global) + attention (regional) & 39.4 & 56.7  & 30.2 & 49.4  & 26.1 & 45.0 \\
5 &\ \ \ \ \ \ dynamic routing (regional) + attention (global) & 39.3 & 56.8  & 30.0 & 49.3  & 26.3 & 45.1 \\
6 &\ \ \ \ \ \ attention (global) + attention (regional)       & 38.9 & 56.3  & 29.8 & 48.8  & 25.9 & 44.7 \\
\hline
7 &\ \ \ \ \ \ w/o context guidance in dynamic routing  & 39.3 & 56.5   & 30.3 & 49.4  & 26.1 & 45.0  \\
\hline
\end{tabular}
\caption{
\label{Table_ablation}
Ablation study of our model on the EN$\Rightarrow$DE translation task. 
\vspace{-10pt}
}
\end{table*}

\textbf{Model Performance} Table \ref{Table_En2DeMainResults} shows the translation quality on  EN$\Rightarrow$DE translation task. It is obvious that DCCN outperforms most of the existing models and all baselines, except Fusion-conv (\textbf{Row} 3) and Trg-mul (\textbf{Row} 4) on METEOR. Note that these two systems are the state-of-the-arts on WMT 2017, with parameter selection based on METEOR. Moreover, we draw two interesting conclusions:

First, DCCN model outperforms Encoder-attention, which uses static source hidden states to attend to visual features. The underlying reasons consist of two aspects: (1) encoder-attention depends on static source representations to extract visual context. By contrast, DCCN utilizes the timestep-specific source-side context vector to extract visual context; and (2) the context-guided dynamic routing mechanism exploits the interactions between different modalities to produce better multimodal context vector. 

Second, although Doubly-attention also uses timestep-specific source-side context vector, DCCN model still achieves a significant improvement over it. This demonstrates again the advantage of modeling the semantic interactions between different modalities on learning multimodal context vectors. 


Finally, following \citeauthor{rnn}~\shortcite{rnn}, 
we divide our test sets into different groups based on the lengths of source sentences, 
and then compare different models in each group.
Figure \ref{Fig_En2De_LengthEffect} reports the BLEU scores of two test sets on different groups. 
Overall, our model still consistently achieves the best performance in most groups. 
Thus, we confirm again the effectiveness and generality of DCCN. 

\begin{figure*}[!t]
\centering
\includegraphics[width=1\linewidth]{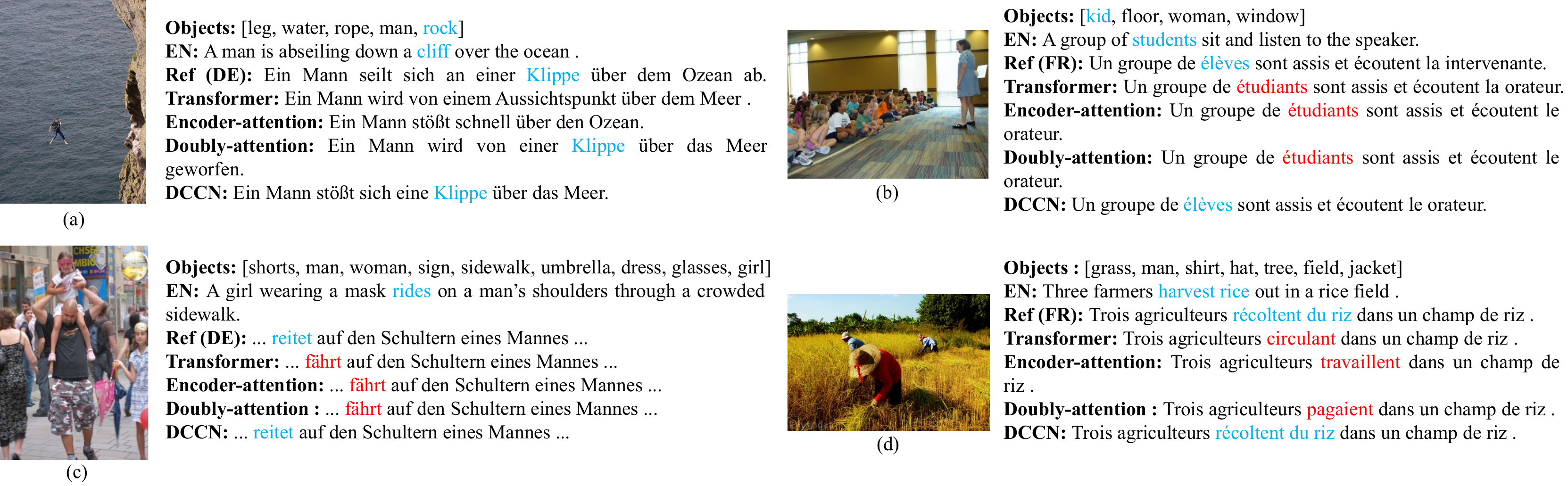}
\caption{
\label{case}
Translation examples of different MMT models. We show the object with the highest probability of each region.}
\end{figure*}

\subsection{Ablation Study}

To explore the effectiveness of different components in DCCN, we further compare our model with the following variants in Table \ref{Table_ablation}: 

(1) \emph{dynamic routing (global)}.
To build this variant, 
we only use global visual features in our model. 
The result in \textbf{Row} 2 indicates that removing the regional visual features lead to performance drop. This result suggests that regional visual features are indeed useful for multimodal representation learning.

(2) \emph{dynamic routing (regional)}.
Unlike the above variant, 
we only use regional visual features to represent the input image in this variant.
According to the result shown in \textbf{Row} 3, 
we observe this change results in a significant performance decline, 
demonstrating that global visual features also bring useful visual information to our model.

\begin{table*}
\centering
\footnotesize
\begin{tabular}{c|l|c|cc|cc|cc}  
\hline
\multirow{3}{*}{\#} & \multirow{3}{*}{\bf Model} & \multirow{3}{*}{\bf\#Params} &\multicolumn{6}{c}{\textbf{EN$\Rightarrow$FR}}\\
\cline{4-9}
& & &\multicolumn{2}{|c|}{\textbf{Test2016}} &\multicolumn{2}{|c|}{\textbf{Test2017}} &\multicolumn{2}{c}{\textbf{MSCOCO}}\\
\cline{4-9}
& & &\textbf{BLEU} & \textbf{METEOR} &\textbf{BLEU} & \textbf{METEOR} \\
\hline
\hline
\multicolumn{9}{c}{\emph{Existing MMT Systems}} \\
\hline
1 & Fusion-conv          \cite{caglayan2017lium} & -- & 53.5 & 70.4  & 51.6 & 68.6  & 43.2 & 63.1 \\
2 & Trg-mul              \cite{caglayan2017lium} & -- & 54.7 & 71.3  & 52.7 & 69.5  & 43.5 & 63.2  \\
3 & Deliberation Network \cite{acl19:twopass}    & -- & 59.8 & 74.4  & --   & --    & -- & --  \\
\hline
\multicolumn{9}{c}{\emph{Our MMT Systems}} \\
\hline
4 &Transformer \cite{vaswani2017attention}           &16.0M  & 60.1 & 75.5  & 53.3 & 69.7  & 44.2 & 64.2   \\
5 &Encoder-attention \cite{Delbrouck:NIPS17workshop} & +1.0M & 60.4 & 75.7  & 53.8 & 69.9  & 44.6 & 64.5    \\
6 &Doubly-attention \cite{helcl2018cuni}             & +3.9M & 60.3 & 75.5  & 53.6 & 70.0  & 44.9 & 64.4   \\
\hline
7 &DCCN & +0.9M  
& $\textbf{61.2}^{\ddagger**\bigtriangleup\bigtriangleup}$  & $\textbf{76.4}^{\ddagger**\bigtriangleup\bigtriangleup}$   
& $\textbf{54.3}^{\ddagger**\bigtriangleup\bigtriangleup}$  & $\textbf{70.3}^{\ddagger*}$ 
& $\textbf{45.4}^{\ddagger**\bigtriangleup}$  &$\textbf{65.0}^{\ddagger**\bigtriangleup\bigtriangleup}$ \\
\hline
\end{tabular}
\caption{
\label{Table_En2Fr}
Experimental results on the EN$\Rightarrow$FR translation task.
\vspace{-10pt}
}
\end{table*}

(3) \emph{dynamic routing $\Rightarrow$ attention}. 
Apparently, one advantage of our model lies in leveraging context-guided dynamic routing to exploit the semantic interactions between different modalities for learning multimodal representation. Here we separately replace the context-guided dynamic routing with the conventional attention mechanism to exploit global visual features, regional visual features, and both of them. Then we investigate the change of model performance. 
To facilitate the following descriptions, we refer to these three variants as 
\emph{dynamic routing (global) + attention (regional)}, 
\emph{dynamic routing (regional) + attention (global)}, 
and \emph{attention (global) + attention (regional)}, respectively. 
From \textbf{Row} 4 to \textbf{Row} 6 of Table \ref{Table_ablation}, 
we observe that \emph{dynamic routing (global) + attention (regional)} slightly outperforms \emph{dynamic routing (global)} while is inferior to DCCN. 
Similarly, the performance of \emph{dynamic routing (regional) + attention (global)} is between \emph{dynamic routing (regional)} and DCCN. 
Moreover, when we use attention mechanism rather than dynamic routing to extract two kinds of visual features, the performance of our model degrades most. 
Based on these experimental results, we can draw the conclusion that context-guided dynamic routing is able to better extract two types of visual features than conventional attention mechanism.

(4)\emph{w/o context guidance in dynamic routing}.
By removing the context guidance from DCCN, we adopt the standard capsule network to extract visual features. 
As shown in \textbf{Row} 7, 
the model performance drops drastically in this case.
This result is consistent with our intuition that the ideal visual features required for translation should be dynamically captured at different timesteps.

\subsection{Case Study}

When encountering ambiguous source words or complicated sentences, it is difficult for MMT models to translate correctly without corresponding visual features. To further demonstrate the effectiveness of DCCN, we display the 1-best translations of the four cases generated by different models, as shown in Figure \ref{case}.

When encountering ambiguous nouns, the regional visual features are more helpful. For example, in case (a), both Transformer and Encoder-attention miss the translation of source word ``\emph{cliff}", while Doubly-attention and DCCN translate it correctly according to the detected object ``\emph{rock}".
In case (b), we can find that the ambiguous source word ``\emph{student}" is translated to ``\emph{étudiants (college students)}" by all baselines, while only DCCN correctly translates it with ``\emph{élèves (generally refers to all students)}" with the help of the detected object ``\emph{kid}". 

When the model has difficulty in translating words out of the predicted objects such as verbs and adjectives, the global visual features are more helpful. In case (c), the source word ``\emph{rides}" is not associated with any object and thus all baselines choose ``\emph{fährt (drive)}", while only DCCN translates it correctly. In case (d), three baselines translate ``\emph{harvest rice}" to ``\emph{circulant (flow)}", ``\emph{travaillent (work)}" and ``\emph{pagaient (paddle)}", respectively. By contrast, only DCCN can produce the correct translation with the help of image information. 

These cases reveal that DCCN can fully utilize complementary visual information to learn more accurate representations and disambiguate during translation in different cases.

\subsection{Results on the EN$\Rightarrow$FR Translation Task}

To investigate the generality of our proposed model, we also conduct experiments on the EN$\Rightarrow$FR translation task. Table \ref{Table_En2Fr} reports the final experimental results. Likewise, no matter which evaluation metric is used, our model still achieves  better performance than all baselines. This result strongly demonstrates again that DCCN is effective and general to different language pairs in MMT.

\section{Conclusion}
In this paper,
we have proposed a novel context-guided capsule network (DCCN) for MMT. As a significant extension of the conventional capsule network, 
DCCN utilizes the timestep-specific source-side context vector to dynamically guide the extraction of visual features at different timesteps, where the semantic interactions between modalities can be fully exploited for MMT via context-guided dynamic routing mechanism. Moreover, we employ DCCN to extract visual features in two complementary granularities: global visual features and regional visual features, respectively. 
Experimental results on English-to-German and English-to-French MMT tasks strongly demonstrate the effectiveness of our model. 
In the future, 
we plan to apply DCCN to other multimodal tasks such as visual question answering and multimodal text summarization.

\section{Acknowledgments}
This work was supported by the Beijing Advanced Innovation Center for Language Resources (No. TYR17002), the National Natural Science Foundation of China (No. 61672440), and the Scientific Research Project of National Language Committee of China (No. YB135-49).

\bibliographystyle{ACM-Reference-Format}
\balance
\bibliography{mm}

\appendix

\end{document}